\documentclass[11pt,a4paper]{article}
\usepackage[hyperref]{emnlp2020}
\usepackage{times}
\usepackage{latexsym}
\usepackage{graphicx}
\usepackage{amsmath}
\usepackage{amssymb}
\usepackage{booktabs}
\usepackage{linguex}

\usepackage{microtype}
\aclfinalcopy
\def\aclpaperid{***}

\newtheorem{myDef}{Definition} 
\newtheorem{myAssumption}{Hypothesis}

\title{Exploring Semantic Capacity of Terms}

\author{Jie Huang$^{*,1,4}$ $\quad$ Zilong Wang$^{*,2}$ $\quad$ Kevin Chen-Chuan Chang$^{1,4}$ \\
 \textbf{Wen-mei Hwu$^{1,4}$ $\quad$ Jinjun Xiong$^{3,4}$} \\
 $^1$University of Illinois at Urbana-Champaign, USA \\
 $^2$University of California at San Diego, USA \\
 $^3$IBM Thomas J. Watson Research Center, USA \\
 $^4$IBM-Illinois Center for Cognitive Computing Systems Research (C3SR), USA \\
 \texttt{\{jeffhj, kcchang, w-hwu\}@illinois.edu} \\
 \texttt{zlwang@ucsd.edu, jinjun@us.ibm.com}
}

\begin{document}
\maketitle
\begin{abstract}
We introduce and study semantic capacity of terms. For example, the semantic capacity of \textit{artificial intelligence} is higher than that of \textit{linear regression} since \textit{artificial intelligence} possesses a broader meaning scope. Understanding semantic capacity of terms will help many downstream tasks in natural language processing. For this purpose, we propose a two-step model to investigate semantic capacity of terms, which takes a large text corpus as input and can evaluate semantic capacity of terms if the text corpus can provide enough co-occurrence information of terms. Extensive experiments in three fields demonstrate the effectiveness and rationality of our model compared with well-designed baselines and human-level evaluations.
\end{abstract}

\newenvironment{starfootnotes}
  {\par\edef\savedfootnotenumber{\number\value{footnote}}
  \renewcommand{\thefootnote}{*} 
  \setcounter{footnote}{0}}
  {\par\setcounter{footnote}{\savedfootnotenumber}}
  
\begin{starfootnotes}
\footnotetext{Asterisk indicates equal contribution. Work done while visiting University of Illinois at Urbana-Champaign.}
\end{starfootnotes}

\section{Introduction}

Terms are not all considered equal. For instance, in computer science, the meaning scope of \textit{artificial intelligence} or \textit{computer architecture} is broader than that of \textit{linear regression}. To study this phenomenon, in this paper, we introduce \textit{Semantic Capacity}, which is a scalar value to characterize the meaning scope of a term. 
A good command of semantic capacity will give us more insight into the granularity of terms and allow us to describe things more precisely, which is a crucial step for downstream tasks such as keyword extraction \citep{hulth2003improved,beliga2015overview,firoozeh2020keyword} and semantic analysis \citep{landauer1998introduction,goddard2011semantic}. 

\begin{figure*}[t]
    \centering
    \includegraphics[width=\linewidth]{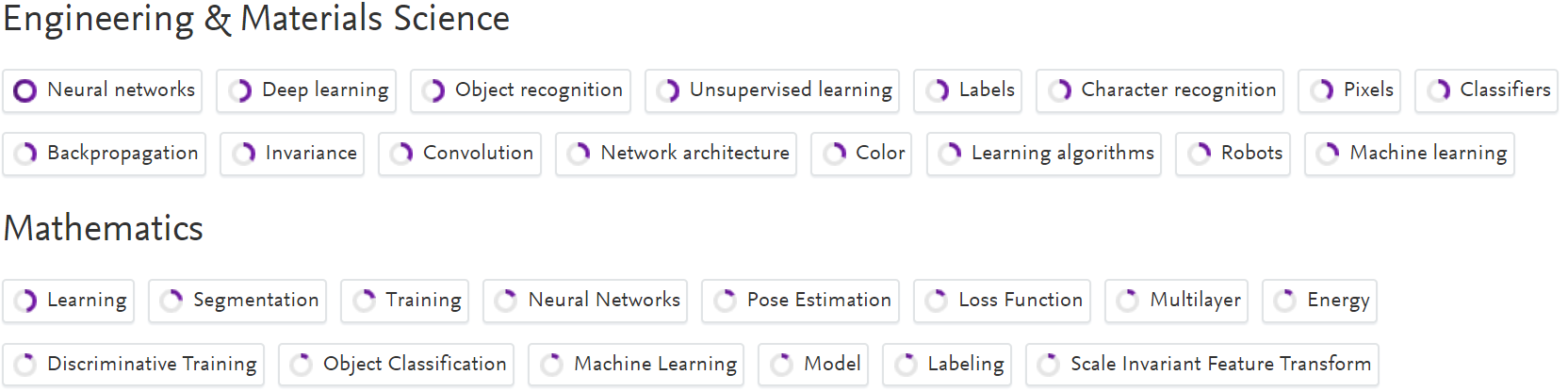}
    \caption{Snapshot of a fingerprint visualization generated by Elsevier Fingerprint Engine.}
    \label{fig:fingerprint}
\end{figure*}

Figure~\ref{fig:fingerprint} shows the fingerprint visualization of a computer scientist, which is generated by Elsevier Fingerprint Engine\footnote{\url{https://www.elsevier.com/solutions/elsevier-fingerprint-engine}}, a popular system that creates an index of weighted terms for research profiling. From the example, we can find that there exist some non-ideal terms, such as \textit{learning} whose semantic capacity is too high, \textit{backpropagation} whose semantic capacity is too low, and even irrelevant terms such as \textit{color}. Understanding semantic capacity of terms will help us to choose better terms to describe entities. Besides, 
combining with other techniques like word similarity, semantic capacity can also help keyword replacement. For instance, 
to describe the computer scientist depicted in Figure~\ref{fig:fingerprint}, if the audience is a layman of computer science, we should use terms with high semantic capacity like \textit{artificial intelligence}. But for an expert in the corresponding domain, we can select terms with low semantic capacity like \textit{object recognition} to make the fingerprint more precise.

However, there are countless terms in human language, which means that it is extremely hard to investigate semantic capacity for all existing terms.
Besides, semantic capacity of terms is also ambiguous in different domains. For instance, \textit{cheminformatics} may be considered as a term with low semantic capacity in computer science and a term with high semantic capacity in chemistry. 

On the other hand, the information on terms we acquire is usually very limited and/or noisy. Although semantic taxonomies such as WordNet \citep{miller1995wordnet} provide rich semantic relations between words, the information is still limited, and these knowledge bases are expensive to maintain and extend. 
Besides, there exists some research work that models hierarchical structures of terms automatically, but most of them suffer from low recall or insufficient precision. For instance, hypernymy discovery \citep{hearst1992automatic,snow2005learning,roller2018hearst} aims at finding \textit{is-a} relations in textual data. 
If we can find all the hypernymy pairs and construct a perfect tree structure that includes every term, the problem of semantic capacity can be solved to some extent. 
However, as far as we know, this is almost impossible in state of the art.

The above analysis shows that we should focus the problem on a specific domain and cover as many terms as possible with easily accessible information. Besides, we should also consider user requirements and deal with terms that are not included at first. 
Therefore, we propose a two-step model that only takes a text corpus as input and can evaluate semantic capacity of terms, provided that the text corpus can give enough co-occurrence signals. Our model consists of the offline construction process and the online query process. The offline construction process measures semantic capacity of terms in a specific semantic space, which narrows the problem to a specific domain and reduces the complexity of the problem to a practical level. The online query process deals with users' queries and evaluates newly added terms that users are interested in. To learn semantic capacity of terms with simple co-occurrences between terms, we introduce the \textit{Semantic Capacity Association Hypothesis} and propose the \textit{Lorentz Model with Normalized Pointwise Mutual Information}, where terms are placed in the hyperbolic space with a novel combination of normalized pointwise mutual information. Finally, norms of embeddings are interpreted as semantic capacity of terms.

The main contributions of our work are summarized as follows:
\begin{itemize}
    \item We study semantic capacity of terms. As far as we know, we are the first to introduce and clarify the definition of semantic capacity.
    \item We propose a two-step model to learn semantic capacity of terms with unsupervised methods. Theoretically, our model can evaluate semantic capacity of any terms appearing in the text corpus as long as the corpus can provide enough co-occurrence signals.
    \item We introduce the \textit{Semantic Capacity Association Hypothesis} and propose the Lorentz model with NPMI, which is a novel application of NPMI to help place terms in the hyperbolic space. We also conceive a novel idea to interpret norms of embeddings as semantic capacity of terms.
    \item We conduct extensive experiments on three scientific domains. Results show that our model can achieve performance comparable to scientific professionals, with a small margin to experts, and much better than laymen.
\end{itemize}

The code and data are available at \url{https://github.com/c3sr/semantic-capacity}.

\section{Methodology}
\label{sec:2}

\begin{figure*}[t]
    \centering
    \includegraphics[width=\linewidth]{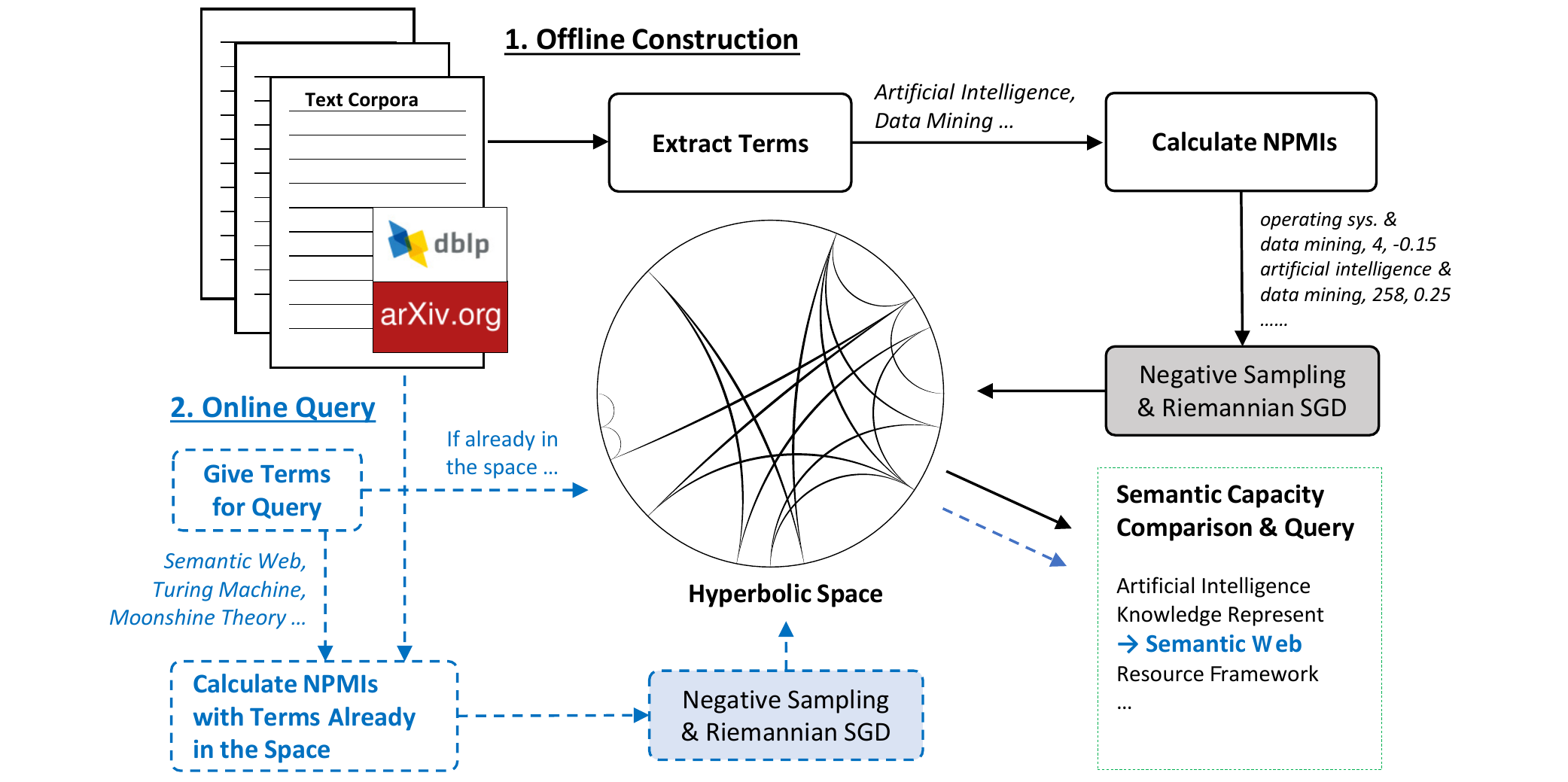}
    \caption{The overview of the two-step model. The model first takes a text corpus as input, and a set of terms are extracted from the corpus. After that, with the training process, terms are placed in the hyperbolic space, and norms of embeddings are interpreted as semantic capacity of terms. For terms that users are interested in but have not already been in the hyperbolic space, the model trains online and returns the corresponding results.
    }
    \label{fig:model}
\end{figure*}

In this section, we introduce the definition of semantic capacity and describe our model in detail. The overview of our model is shown in Figure~\ref{fig:model}.

\subsection{Definition}

The semantic capacity of a term depends on its inherent semantics, the context it is used in, and its associations to other terms in the context. For example, \textit{computer science} is a term with a broad meaning, and it is considered parallel to other terms with broad meanings like \textit{physics} and \textit{materials science}. Besides, \textit{computer science} is also the parent class of some terms with broad meaning scopes like \textit{artificial intelligence} and \textit{computer architecture}.

However, understanding the inherent semantics of terms and modeling the associations between all terms found in human language are impractical due to limited resources. 
Therefore, in this paper, we focus on modeling semantic capacity for terms in a specific domain.
The problem is defined as follows:
\begin{myDef}{(Semantic Capacity)} 
The semantic capacity $SC(\cdot)$ of a term is a scalar value that evaluates the relative semantic scope of a term in a specific domain. And the larger the value, the broader the semantic scope.
\end{myDef}

Semantic capacity reflects the generality of the term in a specific domain of interests, and the larger the value, the more general of such a term.
According to the \textit{Distributional Inclusion Hypotheses (DIH)} \citep{geffet2005distributional}, if $X$ is the superclass of $Y$, then all the syntactic-based features of $Y$ are expected to appear in $X$. Therefore, a term with a broad meaning scope is expected to contain all features of its subclasses, and these subclasses are also expected to contain features of other terms with narrow meaning scopes. Associations between terms can be considered as some kind of syntactic-based features. Therefore, terms with higher semantic capacity are more likely to associate with more terms. 
Besides, in addition to DIH, we also have a new observation that terms like \textit{artificial intelligence} are more likely to have a strong association with its direct subclasses like \textit{machine learning} than descendant classes like \textit{support vector machine}, which means that terms with broader meaning scopes are more likely to associate with terms with broader meaning scopes. Therefore, we propose the \textit{Semantic Capacity Association Hypothesis} as follows:
\begin{myAssumption}{(Semantic Capacity Association Hypothesis)}
Terms with higher semantic capacity will be associated with 1) more terms, and 2) terms with higher semantic capacity than terms with lower semantic capacity. 
\end{myAssumption}

\subsection{Offline Construction Process}

According to the analysis in the introduction, a feasible solution to measure semantic capacity is to focus on a specific domain. Therefore, we first introduce the offline construction process, which aims at learning semantic capacity of terms by taking a large text corpus as input with a number of domain-specific terms extracted from the corpus.

In this paper, to simplify the process and for easier evaluation, we use the public knowledge base Wikipedia Category\footnote{\url{https://en.wikipedia.org/wiki/Wikipedia:Categorization}} as a simple method to extract terms in a specific domain (more details are stated in Section~\ref{sec:datasets}). We can also extract terms from the domain-specific corpus directly by taking some term/phrase extraction methods \citep{velardi2001identification,shang2018automated}. 
After this process, our focus turns to learn semantic capacity of these extracted terms using the text corpus. 

According to the \textit{Semantic Capacity Association Hypothesis}, the key to measuring semantic capacity is to model associations between terms and then put terms in the proper place based on associations among them. 
Specifically, we aim to capture two types of associations between terms: semantic similarity, e.g., the association between AI (artificial intelligence) and ML (machine learning) is stronger than that between AI and DB (database) since ML is closer to AI than DB in meaning; and status similarity, e.g., the association between AI and ML is stronger than that between AI and SVM (support vector machine) since ML is more parallel to AI than SVM.
On the other hand, the number of terms grows exponentially as semantic capacity gets lower, which means we need an exponentially increasing space to place terms. Therefore, we would like to design a method based on associations between terms to place terms in the hyperbolic space where circle circumference and volumes grow exponentially with radius.

Hyperbolic space is a kind of non-Euclidean geometry space represented by the unique, complete, simply connected Riemannian manifold with constant negative curvature. 
Recently, \citet{nickel2017poincare} proposed a hierarchical representation learning model, named the \textit{Poincar\'e ball model}, based on the Riemannian manifold $\mathcal{P}^n = (\mathbf{B}^n, g_p)$, where $\mathbf{B}^n = \{\mathbf{x} \in \mathbb{R}^n: \|\mathbf{x}\|<1\}$ is an open $n$-dimensional unit ball and $g_p$ is the Riemannian metric tensor, which is defined as
\begin{equation}
g_p(\mathbf{x}) = \left( \frac{2}{1-\|\mathbf{x}\|^2} \right)^2 g^E,
\end{equation}
where $\mathbf{x} \in \mathbf{B}^n$ and $g^E$ is the Euclidean metric tensor. The distance function on $\mathcal{P}^n$ is given as
\begin{equation}
d_p(\mathbf{x},\mathbf{y}) \! = \! \cosh^{-1} \! \left(1 \! + \! \frac{2 \cdot \|\mathbf{x}-\mathbf{y}\|^2}{(1-\|\mathbf{x}\|^2)(1-\|\mathbf{y}\|^2)}\right).
\label{eq:dist}
\end{equation}

Given a set of terms and a text corpus, we can count the frequency of co-occurrences $freq(x,y)$ between term $x$ and $y$ by traversing the corpus with a fixed window size. 
We can then learn representations of terms by using co-occurrence information directly based on the Poincar\'e ball model. 
Because of the restriction of hyperbolic space and the distance function, minimizing the loss described in \citep{nickel2017poincare} will be more likely placing terms co-occurring with more terms, especially those with higher co-occurrences, near the center of the Poincar\'e ball.
If co-occurrences capture associations between terms well, according to the \textit{Semantic Capacity Association Hypothesis}, semantic capacity of terms can be interpreted by norms of embeddings to some extent: $SC(x) = 1/\|\mathbf{x}\|$.

However, co-occurrences between terms are very common. There are many valid reasons that terms co-occur. For instance, two terms may co-occur because they are parallel (e.g., \textit{machine learning} and \textit{data mining}), or one term includes the other term (e.g., \textit{artificial intelligence} and \textit{machine learning}). Meanwhile, more generally, irrelevant or distant terms may also co-occur. 
Therefore, the associations modeled by co-occurrences between terms are very noisy, leading to the result that terms with high frequency will co-occur with more terms; thus, they are more likely to be placed near the center.

However, the high frequency of a term cannot guarantee the term's semantic capacity also high. In contrast, there are cases in which terms with less frequency turn out to possess high semantic capacity.
For instance, \textit{theoretical computer science} is a term with high semantic capacity. 
However, it is much less commonly used than its subfield term such as \textit{graph theory}.

With this in mind, to filter noise and better model associations between terms, we introduce \textit{normalized pointwise mutual information (NPMI)} \citep{bouma2009normalized} to help place terms in the hyperbolic space. Letting $\mathcal{W}$ represent the term set, the NPMI value of term $x$ and $y$ is given as
\begin{equation}
npmi(x,y) = - \log \frac{p(x,y)}{p(x)p(y)}/\log p(x,y),
\label{eq:npmi}
\end{equation}
where $p(x,y) = 2 \cdot freq(x,y)/Z$ and $p(x) = freq(x)/Z$ with $freq(x) = \sum_{y \in \mathcal{W}} freq(x,y)$ and $Z = \sum_{x \in \mathcal{W}} freq(x)$.

Compared to pointwise mutual information (PMI), NPMI scales the value between $-1$ and $1$, where $-1$ means $x$ and $y$ never co-occur, $0$ means $x$ and $y$ occur independently, and $1$ means $x$ and $y$ co-occur completely. 
If $x$ and $y$ possess a positive relation, given term $y$, term $x$ will be more likely to occur in the window; thus the NPMI value will be positive.
Therefore, in our model, we set a threshold $\delta>0$ to filter out pairs with negative or weak relations and use the remaining pairs to build the set of associations, which is $\mathcal{D} = \{(x,y): npmi(x,y)>\delta\}$.

According to \citep{nickel2018learning}, the Poincar\'e ball model is not optimal to optimize; therefore, we apply the Lorentz model \citep{nickel2018learning} that can perform Riemannian optimization more efficiently and avoid numerical instabilities. The Lorentz model learns representations in $\mathbf{H}^n = \{ \mathbf{x} \in \mathbb{R}^{n+1}: x_0^2 - \sum_{i=1}^n x_i^2 = 1, x_0 > 0\}$, where the distance function is defined as
\begin{equation}
d_{\ell}(\mathbf{x},\mathbf{y}) = \cosh^{-1}(x_0 y_0 - \sum_{i=1}^n x_i y_i).
\end{equation}

The Lorentz model and the Poincar\'e ball model are equivalent since points in one space can be mapped to the other space \citep{nickel2018learning}. 
Compared to the Lorentz model, the Poincar\'e ball model is more intuitive to interpret the embeddings. Therefore, we adopt the Lorentz model in our training process and use the Poincar\'e ball to interpret semantic capacity of terms.

To learn semantic capacity of terms, we modify the classic loss function of the Lorentz model and propose a new version that considers the strength of association, named \textit{the Lorentz model with NPMI}. 
Letting
\begin{equation}
s(x,y) = \frac{\exp(-d_{\ell}(\mathbf{x},\mathbf{y}))}{\sum_{y' \in \mathcal{N}(x)} \exp(-d_{\ell}(\mathbf{x},\mathbf{y}'))},
\end{equation}
the loss function is given as
\begin{equation}
\mathcal{L}(\Theta) = - \sum_{(x,y) \in \mathcal{D}} npmi(x,y) \cdot \log s(x,y),
\end{equation}
where $\mathcal{N}(x) = \{ y | (x,y) \notin \mathcal{D} \}\cup\{x\} $ is the set of negative examples for $x$, and $\Theta = \{\boldsymbol{\theta}_i\}_{i=1}^{|\mathcal{W}|}$ represents the embeddings of terms, where $\boldsymbol{\theta}_i \in \mathbf{H}^n$. For training, we randomly select a fixed number of negative samples for each associated pair and then try to minimize the distance between points in this pair, against the negative samples. 

Therefore, we aim to solve the optimization problem as
\begin{equation}
\min_{\Theta} \mathcal{L}(\Theta) \quad \text{s.t.}~\forall \boldsymbol{\theta}_i \in \Theta: \boldsymbol{\theta}_i \in \mathbf{H}^n.
\end{equation}

For optimization, we follow \citep{nickel2018learning} and perform Riemannian SGD \citep{bonnabel2013stochastic}.

\subsection{Online Query Process}

Since the terms that we are interested in may not be in the term set $\mathcal{W}$ extracted from the corpus, to evaluate the semantic capacity of newly added terms, we need an online training process to incorporate them into the system.

Assuming a number of terms are already placed in the hyperbolic space, adding a few new terms has little impact on the semantic space and original embeddings. Therefore, we can treat already trained terms as \textit{anchor points} and add new terms into the space dynamically. More specifically, given a new term $a$, we find its co-occurrences with the original terms in $\mathcal{W}$ in the large corpus and calculate the NPMI values for $a$ according to Eq.~\eqref{eq:npmi}. And the optimization problem is then given as
\begin{equation}
\min_{\mathbf{a}} \quad - \sum_{(a,y) \in \mathcal{D}_a} npmi(a,y) \cdot \log s(a,y),
\label{eq:query}
\end{equation}
where $\mathcal{D}_a$ is the set of associations that contain $a$. 

The online query process is illustrated in the blue part of Figure~\ref{fig:model}, where users provide a set of terms. The model first examines whether those terms are already in the space; if so, the system returns the semantic capacity directly. For terms that are not in the space, the system calculates the associations between them and the anchor points in the corpus and solves the optimization problem in Eq.~\eqref{eq:query} by the Lorentz model with NPMI. Finally, semantic capacity of these new terms will be returned as the reciprocal of embedding norms in the Poincar\'e ball. To make the online process more efficient, we can save the statistical information (e.g., co-occurrences with the anchor points) of all terms appearing in the corpus. By doing this, each query can be finished in a short time.

All in all, combining the offline construction and the online query process, we not only deal with the computational problem by focusing on a specific domain, but also have the ability to evaluate semantic capacity of any terms appearing in the text corpus as long as the text corpus can provide enough co-occurrence information. Besides, the online training process can also be considered as a way to extend the semantic space.

\section{Experiments}

In this section, we conduct experiments to validate the effectiveness of our model.

\subsection{Datasets}
\label{sec:datasets}

\begin{table*}[tp!]
    \begin{center}
    \begin{tabular}{c|ccc|ccc}
        \toprule
         & \multicolumn{3}{c|}{number of pairs} & \multicolumn{3}{c}{number of terms} \\
         & all & top 1 & top 2 & all & top 1 & top 2 \\
        \midrule
        Computer Science & 782 & 93 & 325 & 651 & 11 & 109 \\
        \hline
        Physics &  1393 & 105 & 452 & 1090 & 14 & 127 \\
        \hline
        Mathematics & 1070 & 158 & 399 & 826 & 18 & 153 \\
        \bottomrule
    \end{tabular}
    \caption{Statistics of the dataset for $\mathcal{W}_5$.}
    \label{table:dataset}
    \end{center}
\end{table*}

We conduct experiments in three fields, including \textit{computer science}, \textit{physics}, and \textit{mathematics}.

\paragraph*{\textbf{Computer Science}} We use DBLP text corpus\footnote{\url{https://lfs.aminer.cn/misc/dblp.v11.zip}} as input and extract terms from the corpus via Wikipedia Category. More specifically,  we use terms appearing in both the corpus and the top $k$ levels in Wikipedia Category of Computer Science\footnote{\url{https://en.wikipedia.org/wiki/Category:Subfields_of_computer_science}} to build the set of terms $\mathcal{W}_k$, which is considered as a simple term extraction process from the corpus.
Since there are some irrelevant terms (considered as noise) in the category, we filter out terms whose ``Page views in the past 30 days'' $\leq 500$ and length of words $> 3$. Besides, we filter out terms that contain numbers or special symbols. For evaluation, we also extract hypernym-hyponym pairs from Wikipedia Category.

\paragraph*{\textbf{Physics}} We use arXiv Papers Metadata Dataset\footnote{\url{https://www.kaggle.com/tayorm/arxiv-papers-metadata}} as input and extract terms from the corpus via Wikipedia Category of Physics\footnote{\url{https://en.wikipedia.org/wiki/Category:Subfields_of_physics}} in the same way as computer science.

\paragraph*{\textbf{Mathematics}} We also use arXiv Papers Metadata Dataset as input and extract terms from the corpus via Wikipedia Category of Mathematics\footnote{\url{https://en.wikipedia.org/wiki/Category:Fields_of_mathematics}}. Other settings are the same as computer science.

Statistics of the data with respect to $\mathcal{W}_5$ are listed in Table~\ref{table:dataset}. Taking Physics as an example, we extract 1090 terms, including 14 at the top 1 level and 127 at the top 2. Among these
terms, there are 1393 pairs of hypernym-hyponym, including 105 pairs whose hypernym is at the top 1 level and 452 at the top 2.

\subsection{Experimental Setup}

Since our tasks on semantic capacity are brand new and there is no existing baseline that uses co-occurrences between terms to evaluate semantic capacity of terms, we build or adapt the following models for our experiments:
\begin{itemize}
  \item \textbf{Popularity}: A simple method which uses the frequency $freq(\cdot)$ to evaluate the semantic capacity of each term, i.e., $SC(x) \propto freq(x)$.
  \item \textbf{Poincar\'e GloVe}: Poincar\'e GloVe \citep{tifrea2018poincar} is the state-of-the-art model for hierarchical word embedding and hypernymy discovery, which adapts the GloVe algorithm to the hyperbolic space. In our experiments, we use the reciprocal of embedding norms as the semantic capacity of terms.
\end{itemize} 

We also design the following models for ablation study:
\begin{itemize}
  \item \textbf{Euclidean Model (Co-occurrences)}: A variant of our model which uses the Euclidean space instead of the hyperbolic space and models associations between terms by frequency of co-occurrences instead of NPMI.
  \item \textbf{Euclidean Model (NPMI)}: A variant of our model which uses the Euclidean space instead of the hyperbolic space.
  \item \textbf{Lorentz Model (Co-occurrences)}: A variant of our model which models associations between terms by frequency of co-occurrences instead of NPMI.
  \item \textbf{Lorentz Model (NPMI)}: Our model described in Section~\ref{sec:2}.
\end{itemize}

\paragraph*{\textbf{Parameter Settings}}

We performed manual tuning for all models and adopted the following values for the hyperparameters. For all tasks and datasets, to find the co-occurrences between terms, we set window size as 20. For the training of our models, we set embedding size as 20, batch size as 512, number of negative samples as 50, and NPMI threshold $\delta$ as 0.1. We repeated our experiments for 5 random seed initializations. 

All experiments were finished on one single NVIDIA GeForce RTX 2080 GPU under the PyTorch framework.

\begin{table*}[ht]
\small
\begin{center}
\begin{tabular}{c|ccc|ccc|ccc}
\toprule
&\multicolumn{3}{c|}{Computer Science} & \multicolumn{3}{c|}{Physics} & \multicolumn{3}{c}{Mathematics} \\
& all & top 1 & top 2 & all & top 1 & top 2 & all & top 1 & top 2 \\
\midrule 
Popularity & 65.47 & 64.52 & 65.54 & 62.67 & 55.24 & 54.42 & 66.45 & 68.99 & 62.66 \\ 
Poincar\'e GloVe & 65.47 & 70.97 & 67.38 & 61.45 & 56.19 & 54.87 & 63.27 & 68.35 & 64.41 \\ 
\hline 
Euclidean Model (Co-occurrences) & 69.44 & 71.69 & 70.77 & 67.77 & 54.29 & 60.40 & 68.82 & 78.06 & 69.42 \\
Euclidean Model (NPMI) & 71.00 & 73.92 & 75.46 & 58.15 & 47.62 & 53.76 & 64.95 & 65.19 & 65.79 \\
Lorentz Model (Co-occurrences)& 69.57 & 73.12 & 72.00 & 67.34 & 70.48 & 62.39 & 68.66 & 75.95 & 68.92 \\
Lorentz Model (NPMI)& \textbf{74.25} & \textbf{88.39}  & \textbf{77.11} & \textbf{72.52} & \textbf{82.48} & \textbf{74.07} & \textbf{72.34} & \textbf{80.76} & \textbf{73.86} \\
\bottomrule
\end{tabular}
\end{center}
\caption{Results (\%) of semantic capacity comparison tasks.}
\label{table:results1}
\end{table*}

\begin{table*}[ht]
\small
\begin{center}
\begin{tabular}{c|cc|cc|cc}
\toprule
&\multicolumn{2}{c|}{Computer Science} & \multicolumn{2}{c|}{Physics} & \multicolumn{2}{c}{Mathematics} \\
& top 1 & top 2 & top 1 & top 2 & top 1 & top 2 \\
\midrule 
Popularity & 33.84 & 40.32 & 35.94 & 36.06 & 33.02 & 44.27 \\ 
Poincar\'e GloVe & 36.71 & 42.72 & 39.26 & 39.47 & 32.86 & 44.97 \\ 
\hline
Euclidean Model (Co-occurrences) & 28.39 & 40.23 & 40.33 & 44.19 & 30.43 & 45.66 \\
Euclidean Model (NPMI) & 26.62 & 39.12 & 50.66 & 48.45 & 38.63 & 47.10\\
Lorentz Model (Co-occurrences)& 28.11 & 39.17 & 29.51 & 36.53 & 27.24 & 42.62 \\
Lorentz Model (NPMI)& \textbf{18.52} & \textbf{36.57} & \textbf{19.32} & \textbf{34.42} & \textbf{21.90} & \textbf{39.27} \\
\bottomrule
\end{tabular}
\end{center}
\caption{Average rank (\%) of terms at the top 2 levels.}
\label{table:results2}
\end{table*}

\subsection{Evaluation on Offline Construction}

In this section, we test whether the offline construction part of our model can preserve semantic capacity of terms well.
Wikipedia Category can be considered as tree-structured, where each edge is a hypernym-hyponym (broader/narrower) pair so that we can use these pairs for our evaluation.
We first conduct our experiments on the semantic capacity comparison task with term set $\mathcal{W}_5$: given a pair $(x,y)$, determine whether the semantic capacity of $x$ is higher than that of $y$. For each field, we evaluate the accuracy for all pairs (all), pairs with hypernym at the top 1 level (top 1), pairs with hypernym at the top 2 levels (top 2). The results are shown in Table~\ref{table:results1}. 

From the results, we find that the Lorentz model with NPMI outperforms all the baselines significantly, which achieves satisfactory performances in all fields, especially for pairs with hypernym at the top 1 level. Here we should mention that disagreements exist in the evaluation. For instance, in Wikipedia Category, \textit{programming language theory} is the parent class of \textit{programming language} and \textit{computational neuroscience} is the parent class of \textit{artificial intelligence}. 
However, people may also agree that \textit{programming language} is the superclass of \textit{programming language theory} and \textit{artificial intelligence} is broader than \textit{computational neuroscience}. 

\begin{figure}[t]
  \centering
  \includegraphics[width=\linewidth]{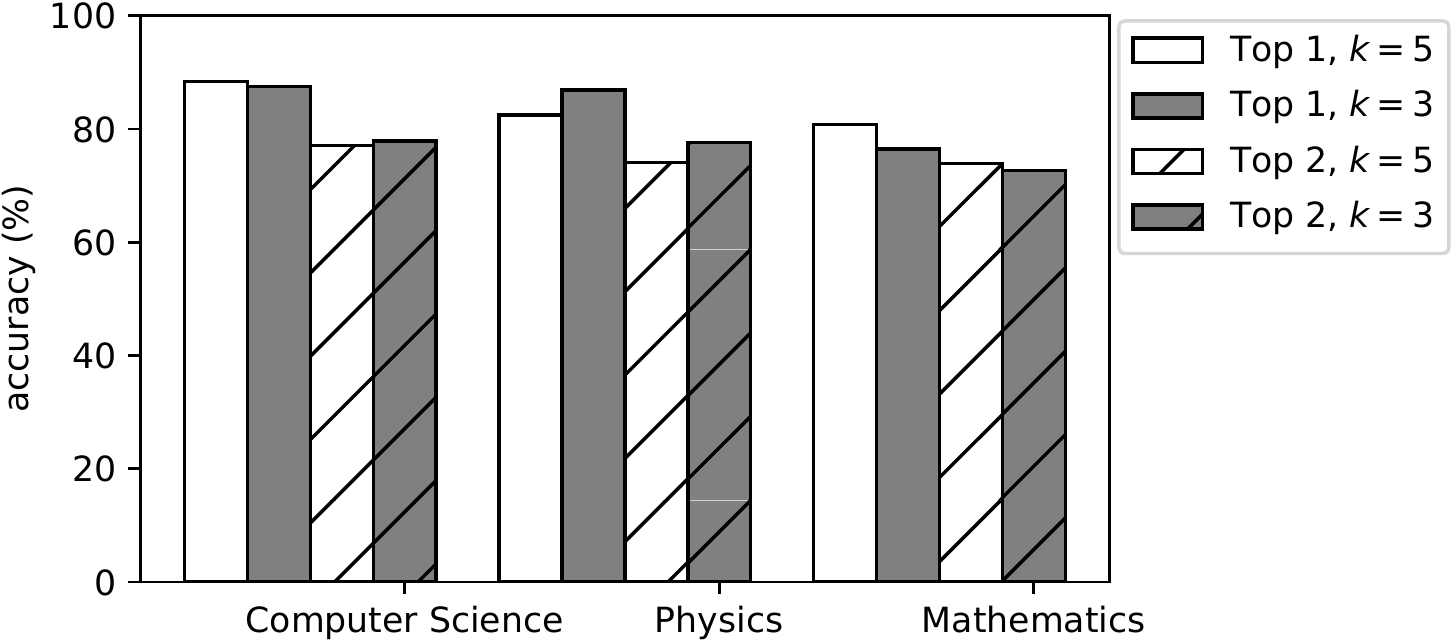}
  \caption{Top 1 and Top 2 accuracies when $\mathcal{W} = \mathcal{W}_k$.}
\label{fig:sensitivity}
\end{figure}

\begin{table*}[ht]
\small
\begin{center}
\begin{tabular}{c|ccc|ccc|ccc}
\toprule
&\multicolumn{3}{c|}{Computer Science} & \multicolumn{3}{c|}{Physics} & \multicolumn{3}{c}{Mathematics} \\
& all & top 1 & top 2 & all & top 1 & top 2 & all & top 1 & top 2 \\
\midrule 
Human Annotation (Layman) & 64.33 & 75.31 & 68.27 & 58.67 & 56.14 & 58.82 & 62.00 & 67.62 & 64.26  \\
Human Annotation (Professional) & 78.33 & 82.72 & 80.32 & 79.67 & 91.23 & 81.96 & 80.00 & \textbf{91.43} & 83.53\\
Human Annotation (Expert) & \textbf{79.33} & 86.42 & 82.73 & \textbf{83.00} & \textbf{94.74} & \textbf{87.06} & \textbf{82.33} & 83.81 & \textbf{84.34} \\
\midrule 
Lorentz Model (NPMI) & 77.40 & \textbf{92.59} & \textbf{84.09} & 78.20 & 91.58 & 79.29 & 76.20 & 80.00 & 79.28 \\
\bottomrule
\end{tabular}
\end{center}
\caption{Results (\%) of semantic capacity query tasks.}
\label{table:query}
\end{table*}

Besides, compared with these variants of our model, the Lorentz model with NPMI has a significant performance improvement over them, which indicates the effectiveness of using filtered NPMI to characterize associations between terms and shows the superiority of placing terms in the hyperbolic space. In terms of training speed, taking the offline construction in computer science as an example, compared with the run time of the Lorentz model with co-occurrences (51s), the Lorentz model with NPMI also has an improvement in efficiency (30s).

To compare with methods based on lexico-syntactic patterns, we also try Hearst patterns (with extended patterns) \citep{hearst1992automatic} to find the hypernymy relations for physical terms. The result shows that only 2.5\% (35/1393) of the hypernymy pairs are detected, i.e., almost impossible to measure semantic capacity of terms. 

In addition to evaluating on the pairs, we introduce a metric to evaluate the performance in a different way. 
Since semantic capacity is not strictly divided by levels of terms, it is possible that the semantic capacity of a term at the higher level is lower than that of a term at the lower level. But in general, the average rank of terms at the higher level should be higher than that of terms at the lower level. Therefore, we use the average rank of terms at the top $k$ levels $(AR_k)$ as a metric to evaluate the performance, which is defined as 
\begin{equation}
AR_k = \frac{1}{|\mathcal{W}_k|} \sum_{x \in \mathcal{W}_k} \frac{rank(x)}{|\mathcal{W}|},
\end{equation}
where $|\mathcal{W}|$ denotes the cardinality of the term set and $rank(x)$ is the ranking (being the top rank or the highest semantic capacity) of term $x$ evaluated by the model. In other words, when $k$ is small, the smaller $AR_k$, the better. For terms at the top $1$ level, the metric is sensitive to misordered terms, and the value will grow a lot when a term is ranked low. Again, semantic capacity is not strictly divided by levels of terms, but in general, terms at the higher level should have higher ranks (smaller in value). 
Results in Table~\ref{table:results2} show that our model achieves the best performance, and the results are consistent with the results of the semantic comparison task.

\paragraph*{\textbf{Sensitivity to Term Set}} 

The training process is affected by the term extraction process. 
Therefore, we want to detect model sensitivity with respect to the term set. 
For this purpose, we use $\mathcal{W}_5$ and $\mathcal{W}_3$ in each field as the term set respectively and conduct the semantic capacity comparison task for pairs with hypernym at the top 1 level and pairs with hypernym at the top 2 levels.

From Figure~\ref{fig:sensitivity}, we can see the results are relatively stable. On the one hand, compared to $\mathcal{W}_3$, $\mathcal{W}_5$ contains more terms, which means term set $\mathcal{W}_5$ is more complete, but the training time also increases with the number of terms. On the other hand, since noise increases with the level in Wikipedia Category, $\mathcal{W}_5$ contains more noisy terms than $\mathcal{W}_3$.
In short, how to choose the term set depends on many factors, such as the task we care about and the noise contained in the term set we acquire.

\subsection{Evaluation on Online Query}

In this section, experiments are conducted to validate the performance of the online query process on evaluating semantic capacity of newly added terms.
We randomly select 100 hypernym-hyponym pairs at the top 3 levels of each evaluation set for online query and use the remaining terms in $\mathcal{W}_3$ for offline construction.
We compare our model with human annotation by three groups of people, where each pair is labeled by three unique people. Details of human annotation are listed as follows:
\begin{itemize}
    \item \textbf{Human Annotation (Layman)}: Human annotation by workers on Amazon Mechanical Turk\footnote{\url{https://www.mturk.com}} with ``HIT Approval Rate'' $\geq$ 95\% (considered as high quality).
    \item \textbf{Human Annotation (Professional)}: Human annotation by non-major students in the United States. 
    Specifically, we ask math, computer science, physics students to conduct annotation tasks for physics, math, computer science, respectively.
    \item \textbf{Human Annotation (Expert)}: Human annotation by corresponding major students.
\end{itemize}

From the results shown in Table~\ref{table:query}, we find that our model far outperforms human annotation by laymen in all fields. And the performance of our model is comparable to that of human annotation by professionals, with a small margin to experts. The results also imply disagreements exist in the evaluation since experts cannot achieve accuracies close to 100\%. 
Besides, for both our model and human annotation, the top 1 accuracy is usually higher than top 2 accuracy, and the top 2 accuracy is higher than accuracy for all pairs, which is in line with common sense that semantic capacity of terms at the top levels is usually easier to evaluate.
In short, the results demonstrate the effectiveness of our model for evaluating semantic capacity of newly added terms. Furthermore, our model can be applied to semantic capacity query for terms that are not included in the offline process.

\section{Related Work}

Our work is related to research on lexical semantics \citep{cruse1986lexical}. Among them, hypernymy, also known as \textit{is-a} relation, has been studied for a long time. A well-known method is the Hearst patterns \citep{hearst1992automatic}, which extracts hypernymy pairs from a text corpus by hand-crafted lexico-syntactic patterns. Inspired by the Hearst patterns, some other pattern-based-methods like \citep{snow2005learning,roller2018hearst} are proposed successively. On the other hand, hypernymy discovery based on distributional approaches has also attracted widespread interest \citep{weeds2004characterising,lenci2012identifying,chang2018distributional}.

The techniques our model based on are related to research on learning representations of symbolic data in the hyperbolic space \citep{krioukov2010hyperbolic,nickel2017poincare,nickel2018learning}. 
Since text preserves natural hierarchical structures, \citet{dhingra2018embedding} design a framework that learns word and sentence embeddings in an unsupervised manner from text corpora, \citet{tifrea2018poincar} propose Poincar\'e GloVe to learn word embeddings based on the GloVe algorithm in the hyperbolic space, \citet{aly2019every} use Poincar\'e embeddings to improve exiting methods to domain-specific taxonomy induction, and \citet{le2019inferring} propose a method to predict missing hypernymy relations and correct wrong extractions for Hearst patterns based on the hyperbolic entailment cones \citep{ganea2018hyperbolic}.

\section{Conclusion}

In this paper, we explore semantic capacity of terms. 
We first introduce the definition of semantic capacity and propose the \textit{Semantic Capacity Association Hypothesis}.
After that, we propose a two-step model to investigate semantic capacity of terms, which consists of the offline construction and the online query processes. The offline construction process places domain-specific terms in the hyperbolic space by our proposed Lorentz model with NPMI, and the online query process deals with user requirements, where semantic capacity is interpreted by norms of embeddings. Extensive experiments with datasets from three fields demonstrate the effectiveness and rationality of our model compared with well-designed baselines and human-level evaluations.

In addition, while semantic capacity studied in this paper is restricted to a specific domain, we believe the notion of semantic capacity can be extended to all terms in human language. 
The extension of the scope will be the future work.

\section*{Acknowledgments}

This material is based upon work supported by the National Science Foundation IIS 16-19302 and IIS 16-33755, Zhejiang University ZJU Research 083650, UIUC OVCR CCIL Planning Grant 434S34, UIUC CSBS Small Grant 434C8U, and IBM-Illinois Center for Cognitive Computing Systems Research (C3SR) - a research collaboration as part of the IBM Cognitive Horizon Network. Any opinions, findings, and conclusions or recommendations expressed in this publication are those of the author(s) and do not necessarily reflect the views of the funding agencies. We thank the anonymous reviewers for their valuable comments and suggestions.

\bibliographystyle{acl_natbib}
\bibliography{main}

\end{document}